\documentclass[letterpaper, 10 pt, conference]{ieeeconf} 

\IEEEoverridecommandlockouts
\usepackage{graphics} 
\usepackage{epsfig} 

\usepackage{cite} 
\usepackage{amsmath}
\usepackage{amssymb} 
\usepackage{subcaption}
\usepackage{graphicx}
\usepackage{todonotes}
\usepackage{makecell}
\usepackage[pdfa]{hyperref}
\usepackage{comment}
\usepackage{multirow}
\usepackage{ctable}
\usepackage{makecell}
\usepackage{multirow}
\usepackage{multicol}
\usepackage{soulutf8}

\usepackage[bottom]{footmisc} 

\usepackage[ruled,vlined]{algorithm2e}
\usepackage{algpseudocode}

\usepackage[normalem]{ulem}
\useunder{\uline}{\ul}{}

\usepackage{float}
\usepackage{dblfloatfix} 

\usepackage{xsavebox} 
\usepackage{atbegshi} 

\xsavebox{PageBGPicture}{
	\begin{tikzpicture}
	\node [rectangle, minimum width=15cm, minimum height=1.5cm, align=center, text = gray] (box) {This paper has been accepted for publication at IEEE Robotics and Automation Letters (RA-L) 2022. \\ \copyright 2022 IEEE. Personal use of this material is permitted. Permission from IEEE must be obtained for all other uses, in any current or future \\ media,  including reprinting/republishing  this material for advertising or promotional purposes, creating new collective works, for resale or \\ redistribution to servers or lists, or reuse of any copyrighted component of this work in other works.};
	\end{tikzpicture}
}

\AtBeginShipout{
	\AtBeginShipoutUpperLeft{\raisebox{-\height}{\xusebox{PageBGPicture}}}
}

\graphicspath{ {./img/} }
\newcommand\scalemath[2]{\scalebox{#1}{\mbox{\ensuremath{\displaystyle #2}}}}

\title{\LARGE \bf Free as a Bird:  Event-based Dynamic Sense-and-Avoid \\ for Ornithopter Robot Flight
}

\author{J.P. Rodr\'{\i}guez-G\'omez, R. Tapia, M.M. Guzmán, J.R. Mart\'{\i}nez-de Dios and A. Ollero%
\thanks{
The authors thank Jesus Tormo for his help with the robot electronics, and Angela Romero for her support on the validation experiments.
This work was performed within GRIFFIN ERC Advanced Grant (Action 788247) (\url{https://griffin-erc-advanced-grant.eu}) funded by the European Research Council. Partial funding was obtained from European Commission H2020 AERIAL-CORE project (Grant H2020-2019-871479) and from Plan Estatal de Investigación Científica y Técnica y de Innovación of the Ministerio de Universidades del Gobierno de España (FPU19/04692).\newline
The authors are with the GRVC Robotics Lab Sevilla. Universidad de Sevilla, Spain %
{\tt email: \{jrodriguezg, raultapia, mguzmang, jdedios, aollero\}@us.es}}%
}

\begin{document}
\maketitle
\thispagestyle{empty}
\pagestyle{empty}

\begin{abstract}
Autonomous flight of flapping-wing robots is a major challenge for robot perception. Most of the previous \textit{sense-and-avoid} works have studied the problem of obstacle avoidance for flapping-wing robots considering only static obstacles. This paper presents a fully onboard dynamic \textit{sense-and-avoid} scheme for large-scale ornithopters using event cameras. These sensors trigger pixel information due to changes of illumination in the scene such as those produced by dynamic objects. The method performs \textit{event-by-event} processing in low-cost hardware such as those onboard small aerial vehicles. The proposed scheme detects obstacles and evaluates possible collisions with the robot body. The onboard controller actuates over the horizontal and vertical tail deflections to execute the avoidance maneuver. The scheme is validated in both indoor and outdoor scenarios using obstacles of different shapes and sizes. To the best of the authors' knowledge, this is the first event-based method for dynamic obstacle avoidance in a flapping-wing robot.
\end{abstract}


\section{Introduction}
\label{sec:intro}
Flapping-wing robots, also known as ornithopters, have recently attracted significant R\&D  interest. They can perform agile maneuvers \cite{zufferey2021design} and combine flapping and gliding modes to reduce energy consumption \cite{de2020flapping}. Besides, flapping-wing robots are often made of soft materials making them less dangerous than multirotors in case of collision \cite{haider2020recent}. Flapping-wing flight describes novel perception challenges different from those in multirotor flight. First, ornithopters generate lift and thrust by flapping strokes, causing mechanical vibrations and wide abrupt movements that highly impact onboard perception \cite{gomezeguiluz2019towards}. Besides, they have strict payload and energy limitations, which strongly constrain the installation of sensors and additional hardware, involving strict limitations on the onboard processing capacity. In fact, most reported ornithopter perception and control methods, e.g., \cite{maldonado2020adaptive, farrell2018review}, are executed offboard and use measurements from external sensors such as motion capture systems. 
We are interested in autonomous navigation of ornithopter robots, and particularly in avoidance of dynamic obstacles. While static obstacles can often be assumed within the map (and addressed through trajectory planning) or detected with additional sensors such as LiDAR, this work deals with the avoidance of unexpected dynamic obstacles in ornithopters, which require fast onboard detection and avoidance in contrast to the strict payload and resource constraints of these platforms. The perception scheme is based on event cameras. They are robust to motion blur and lighting conditions and have moderate weight and low energy consumption. Hence, they are suitable to deal with the flapping-wing flight perception challenges \cite{gomezeguiluz2019towards}. Besides, event cameras are suitable for dynamic obstacle detection by directly providing pixel information of moving objects in the scene. Additionally, efficient event-based processing techniques can provide estimates at very high rates. Many successful event-based perception techniques have been developed \cite{gallego2019event}. 
\begin{figure}[t]
    \centering
    \includegraphics[trim={9cm 5cm 14cm 12cm},clip,width=0.99\linewidth]{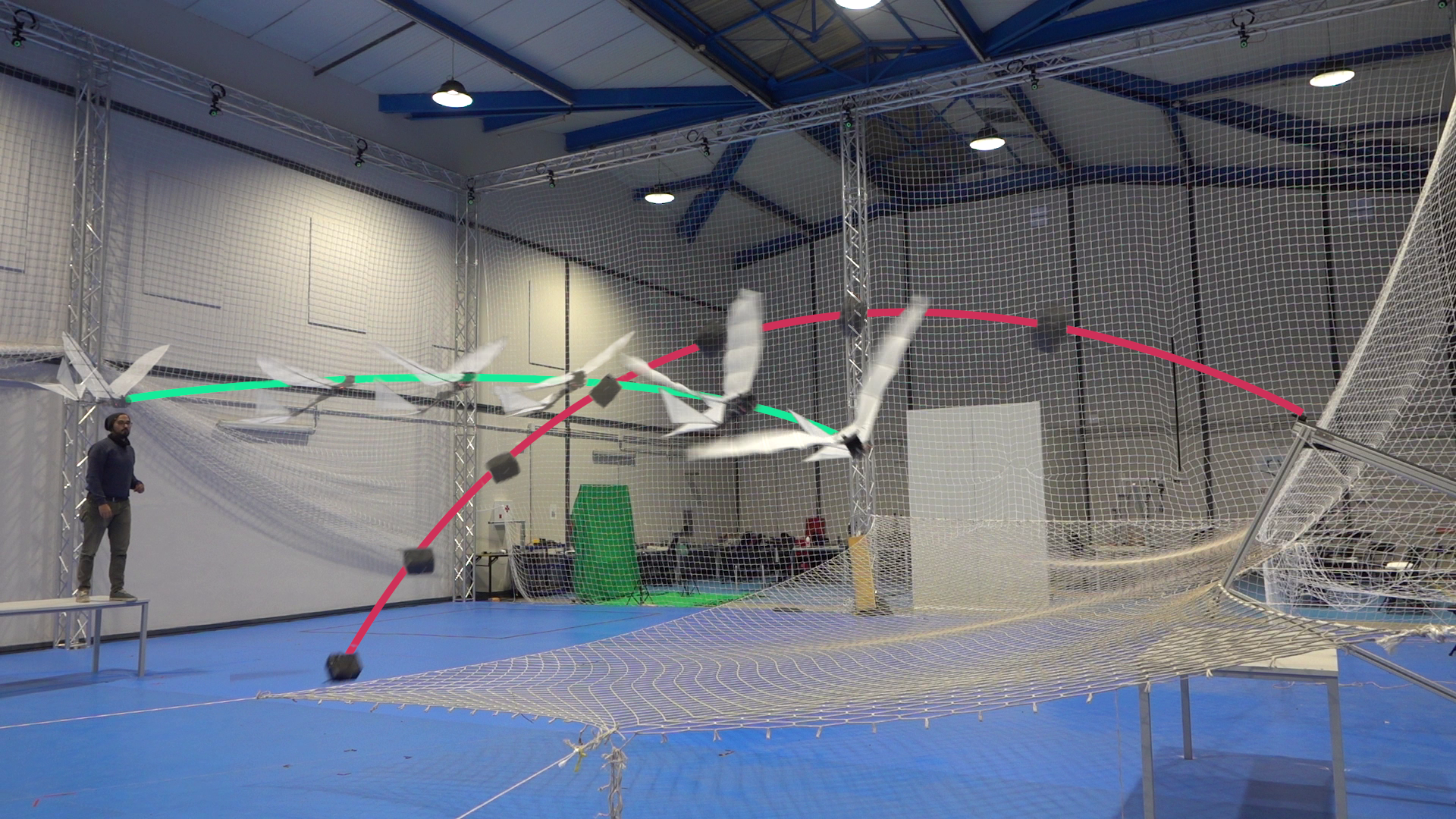}
    \caption{Image sequence of the \textit{E-Flap} robot performing an obstacle avoidance maneuver in an experiment.}
    \label{fig:intro}
    \vspace{-0.5cm}
\end{figure}

This paper presents a dynamic obstacle \textit{sense-and-avoid} method for ornithopter robots. By processing only one onboard event camera, the robot rapidly detects dynamic obstacles and modifies its trajectory to avoid them, exploiting the low latency of event cameras and the agility of ornithopters’ flight. Dynamic obstacles are segmented using the spatio-temporal event information from objects that move with a different velocity than the background. An event optical flow method estimates its direction, and a reactive evasive maneuver strategy rapidly evaluates and prevents collisions. The method is implemented for online execution in resource-constrained hardware and is evaluated in the GRIFFIN \textit{E-Flap} large-scale ornithopter \cite{zufferey2021design}, see Fig. \ref{fig:intro}, in indoor and outdoor experiments. To the best of the authors' knowledge, this is the first event-based obstacle avoidance method designed for and validated in flapping-wing robots.

The main contributions of the paper are: (1) an event-based dynamic object motion estimation method designed to perform in low-resource hardware; (2) a reactive obstacle avoidance method for large-scale ornithopters providing low latency onboard perception and control; and (3) experimental validation indoors and outdoors on a large-scale ornithopter. 
This paper is organized as follows. Section \ref{sec:SoA} briefly summarizes the main works in the topics addressed in the paper. The general diagram of the event-based ornithopter dynamic obstacle avoidance scheme and its main components are described in Sections \ref{sec:problem} and \ref{sec:methods}. Section \ref{sec:experimental} presents the experimental validation and robustness analyses. Section \ref{sec:conclusion} closes the paper and highlights the main future steps.

\section{Related work}
\label{sec:SoA}

\begin{table}[b!]
\scalebox{0.94}{
\renewcommand{\tabcolsep}{1.5pt}
\renewcommand{\arraystretch}{1.2}
\begin{tabular}{c | c c c c c c c}
 \cline{2-8}
& \makecell{\textbf{Flight} \\ \textbf{speeds}} & \makecell{\textbf{Proc.} \\ \textbf{ Weight}} & 
\makecell{\textbf{\# of}\\ \textbf{cam}} &   \makecell{\textbf{C/GPU}\\ \textbf{cores}} &\textbf{Proc.} & \textbf{Accum.} & \makecell{\textbf{Proc.} \\ \textbf{Time}} \\ \hline 
\cite{mueggler2015towardsevasive} &  $\sim$0 m/s &  $\sim$49 g & 2 &  4/0 & \textit{e-by-e} & Async & --.\\
\cite{Sanket2019EVDodgeEA} &  $\sim$0 m/s & $\sim$150 g & 2 & 6/1 & \textit{e-images} & 30 ms & 12 ms/\textit{e-images}  \\
\cite{falanga2020dynamic} & [0,1.5] m/s &  $\sim$150 g &  1-2 & 6/1 & \textit{e-images} & 10 ms & 3.56 ms/\textit{e-images}  \\
\textbf{Ours} & [2,4] m/s & 28.5 g & 1 & 6/0 & \textit{e-by-e} & Async. & 4 ms/package \\
\specialrule{.1em}{.05em}{.05em}
\end{tabular}
}
\caption{Comparison of our approach with other dynamic obstacle avoidance methods --for quadrotors. \textit{e-by-e} regards to \textit{event-by-event}, while \textit{e-images}, to \textit{event images}.}
\label{table:soa}
\end{table}

Reactive obstacle avoidance focuses on generating avoidance robot actions without relying on globally consistent map information. It can be categorized into \textit{map-based} and \textit{map-less} approaches \cite{hrabar2011reactive}.  The first builds a local map to compute obstacle-free trajectories \cite{oleynikova2015reactive}, while the second, aims at detecting nearby obstacles and directly performs the avoidance action \cite{mueggler2015towardsevasive}. \textit{Map-less} methods provide faster obstacle avoidance and are suitable for platforms with limited processing capacity. 
A work analysing the perception latency in a high-speed \textit{sense-and-avoid} scenario with a quadrotor is presented in \cite{falanga2019howfast}. The work in \cite{ma2018saliency} presents a \textit{map-less} obstacle avoidance solution to avoid flying obstacles using a saliency-based reinforcement learning approach. 

Recent advances in ornithopter development have led to the necessity of developing onboard perception methods capable of providing information for navigation, landing, and perching. 
Optical flow estimation onboard a Micro Aerial Vehicle (MAVs) flapping-wing robot is presented in \cite{bermudez2009optical}. The method sub-samples input images and uses a motion detection algorithm to compute the optical flow. 
Authors in \cite{deCroon2010appearance} use the object appearance variation and optical flow to perform obstacle avoidance with a monocular camera. Obstacle detection and avoidance are computed in a ground station due to the weight limitation of the platform. 
A stereo-vision obstacle avoidance strategy for small-scale ornithopters is presented in \cite{tijmons2017obstacle}. The method computes sparse disparity maps from points with relatively high certainty for obstacle depth estimation. Obstacle avoidance is achieved through the droplet strategy defining the necessary obstacle-free area in front of the robot to guarantee safe avoidance maneuvers. 
The advantages offered by event cameras have increased the research interest in computer vision and robotics communities \cite{gallego2019event}.  
Their temporal $\mu$s resolution and high dynamic range motivate their use for aerial robot perception. An event-based optical flow approach for autonomous MAV landing using a downwards orientated event camera is presented in \cite{pijnacker2018vertical}.
The event-based line tracker in \cite{gomezeguiluz2020async} provides fast and stable references for quadrotor visual servoing to perform bionspired landing trajectories. The drone racing dataset \cite{delmerico2019we} including event data intends to encourage the development of perception methods for high-speed drone maneuvers. The work in \cite{sun2021autonomous} accumulates events to build \textit{event images} in order to estimate the position and orientation of a quadrotor using a visual odometry method and closing the loop for autonomous flight subject to rotor failures. In \cite{rodriguez2021autotuned}, an auto-tuned event-based vision scheme performs intruder detection on board an autonomous quadrotor in surveillance missions.

Recently, some event-based methods for reactive obstacle avoidance for quadrotors have been presented. Work \cite{mueggler2015towardsevasive} describes an evasive method for dynamic obstacles using stereo event cameras on a quadrotor. The object trajectory is computed and propagated in time to predict collisions. An event-based dodging system for quadrotors is presented in \cite{Sanket2019EVDodgeEA}. It performs image deblurring, odometry estimation, and moving object segmentation using Deep Learning. 
Work \cite{falanga2020dynamic} presents a dynamic obstacle avoidance method for quadrotors that relays in event motion compensation to detect events triggered by moving obstacles, and uses a potential field approach to execute the evasive maneuver. These methods were designed for quadrotors and are not suitable for ornithopters due to the  strong differences between both types of platforms. 

Table \ref{table:soa} compares our approach and the methods in \cite{mueggler2015towardsevasive,Sanket2019EVDodgeEA,falanga2020dynamic}. First, none of these methods is designed for platforms that move at medium-high velocities. They were validated in quadrotors hovering (or moving up to $1.5$ m/s in the case of \cite{falanga2020dynamic}), in which the majority of the triggered events are caused by the obstacle's motion --simplifying obstacle detection. Our method has been designed for flapping-wing robots which require a minimum flight speed of $3$ m/s to flight \cite{zufferey2021design} and suffer from vibrations and angular and linear velocities \cite{gomezeguiluz2019towards}, which trigger additional events caused by the static background, requiring specific moving obstacle detection methods. Moreover, methods \cite{mueggler2015towardsevasive,Sanket2019EVDodgeEA,falanga2020dynamic} require significantly high onboard weight and computational resources, which could not be mounted on a large-scale ornithopter. Methods \cite{Sanket2019EVDodgeEA} and \cite{falanga2020dynamic} use a powerful embedded computer, a flight board, and an autopilot. Besides, the three methods use two event cameras, which also increases their computational requirements. Although \cite{falanga2020dynamic} includes a solution with a monocular camera, it uses an additional board to run  vision-based state estimation. Conversely, our method uses one only event camera and runs in a single lightweight onboard computer with low computational capacity to satisfy the ornithopter payload restrictions while providing a fast response (control loop closing at 250 Hz) to deal with the high flight velocities. Finally, \cite{Sanket2019EVDodgeEA} and \cite{falanga2020dynamic} segment moving objects by processing \textit{event images} resulting from accumulating the incoming events. Hence, on these methods event processing starts after the events have been accumulated, resulting in delays between event generation and processing, even in cases in which \textit{event image} processing times are lower than the \textit{event image} accumulation times, such as in \cite{falanga2020dynamic}. Conversely, our method adopts \textit{event-by-event} processing exploiting the asynchronous nature of event generation and enabling shorter obstacle detection times, which is interesting in our case due to the medium-high ornithopter flight velocities. 

\section{General description}
\label{sec:problem}

\textit{Sense-and-avoid} of agile aerial robots such as ornithopters requires low latency perception for fast obstacle detection. Event cameras provide visual information with $\mu$s resolution triggered by changes of illumination. These sensors have a high dynamic range ($\sim$120 dB) and are robust to illumination conditions and to motion blur, typically desired features in aerial robotics \cite{rodriguez2021autotuned}. Previous works have explored the advantages of event cameras in ornithopters \cite{rodriguez2021griffin} \cite{gomez2021why} and agile quadrotors \cite{delmerico2019we}. Event cameras are compact, have moderate weight, and report low-energy consumption. 

The development and integration of \textit{sense-and-avoid} systems for ornithopters entail additional requirements to those considered in other UAVs such as multirotors. First, ornithopters have strict payload capacity and space restrictions which limit the installation of powerful processing hardware, mechanical stabilizers (e.g., gimbals), and sensors. Thus, onboard hardware is carefully selected to satisfy payload and weight balance restrictions and enable real-time onboard processing. Moreover, flapping-wing robots present complex kinematics and dynamics. They are non-holonomic robots using few actuators to control their position and orientation. 
They generate lift and thrust by flapping their wings, also producing forward, backward, or lateral movements.  
These aspects set additional requirements for obstacle avoidance including low latency obstacle detection and  evasion strategies that consider the platform kinematics and dynamics. 

\begin{figure}[t]
	\centering
	{\includegraphics[trim={2cm 0 0 0},clip,width=0.95\linewidth]{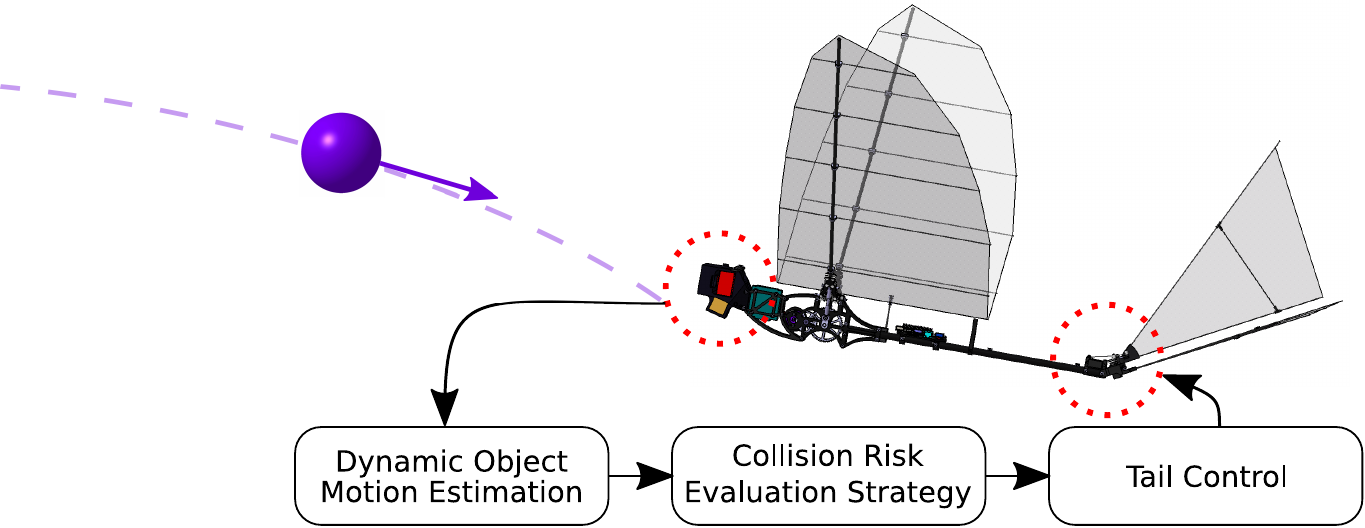}}
	\caption{General diagram of the proposed event-based \textit{sense-and-avoid} scheme for flapping-wing flight.}
	\label{fig:problem description}	
	\vspace{-0.5cm}
\end{figure}

The general diagram of the proposed \textit{sense-and-avoid} scheme is shown in Fig. \ref{fig:problem description}. The \textit{Dynamic Obstacle Motion Estimation} method, see Section \ref{subsec:obstacle_detection}, detects dynamic obstacles and estimates their motion in the image plane. It uses the spatio-temporal information of events triggered by dynamic objects moving with a different velocity than the background. Although many motion-based segmentation methods using traditional framed cameras have been proposed \cite{zhang2001segmentation}, the use of event-based vision has interesting advantages in our problem. First, event-based processing provides natural robustness against motion blur and changes in lighting conditions. Further, motion segmentation using framed cameras process full frames, while event-based methods only analyse asynchronous events enabling faster processing, $250$ Hz in the experiments shown in Section \ref{sec:experimental}.
The \textit{Collision Risk Evaluation Strategy} module evaluates the risk of collision considering the robot geometry, see Section \ref{subsec:evasive_policy}. If a collision risk situation is detected an avoidance maneuver is performed by the ornithopter. The optical flow of detected obstacles is used to reactively command the ornithopter to avoid collisions. The adopted \textit{Tail Control} method meets both robustness and simplicity requirements. It actuates on the vertical and horizontal tail deflections to change the ornithopter flight direction. The value of the control commands is adjusted using a simplification of the robot model to consider the dynamic constraints enclosed in the evasion maneuver.

The event processing method leverages ASAP \cite{tapia2020asap}, which adapts event packaging such that events are processed as soon as possible while avoiding computational overflows, and ensures control closing at 250 Hz with onboard resource-constrained hardware. 

\section{Methods}
\label{sec:methods}

\subsection{Dynamic Object Motion Estimation}
\label{subsec:obstacle_detection}

\vspace{-0.1cm}
\begin{figure*}[b]
\begin{center}
\begin{minipage}{0.63\linewidth}
\renewcommand{\tabcolsep}{2.6pt}
\renewcommand{\arraystretch}{1.2}
\begin{center}
\includegraphics[width=0.95\textwidth ]{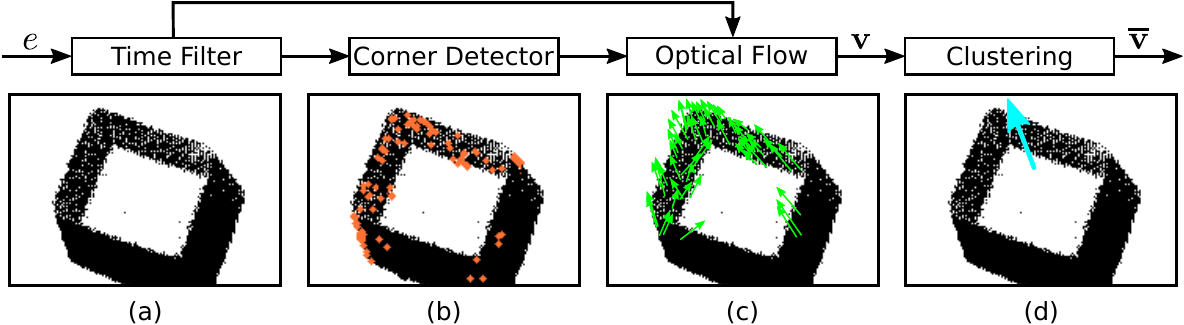}
\captionof{figure}{Block diagram of the \textit{Dynamic Object Motion Estimation} method: a) \textit{Time Filter} detects events triggered by moving objects; b) \textit{Corner Detector} finds relevant features of the object; c) \textit{Optical Flow} estimates the motion of events in the image plane; and d) \textit{Clustering} estimates the object flow using the flow of events belonging to the dynamic object. For clearer visualization the results are shown in \textit{event images} accumulating events every $10$ ms.}
\label{fig:obstacle_detection_diagram}
\end{center}
\end{minipage}
\hspace{1pt}
\begin{minipage}{0.35\linewidth}
\includegraphics[width=0.49\linewidth]{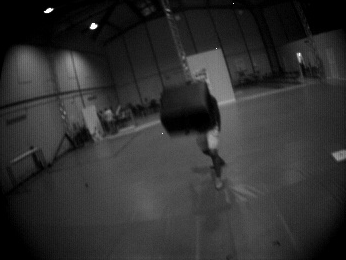}
\includegraphics[width=0.49\linewidth]{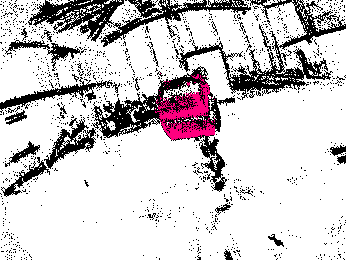}
\captionof{figure}{Event-based dynamic object detection: right) frame from the APS sensor of the DAVIS 346; left) \textit{event image} of events accumulated during $10$ ms (in black) including events detected to belong to the moving object (magenta).}
\label{fig:moving-object-detection}
\end{minipage}
\end{center}
\vspace{-2.5em}
\end{figure*}

The proposed method performs \textit{event-by-event} processing to exploit the asynchronous nature of event cameras. 
Each event is defined by the tuple $e = (\mathbf{x}, ts, p)$, where $\mathbf{x}$ represents the pixel coordinates $(u,v)$, $ts$ is the timestamp of the event and $p$ is the polarity either positive or negative. The block diagram of the proposed method is shown in Fig. \ref{fig:obstacle_detection_diagram}. The \textit{Time Filter} module detects events belonging to moving objects using as reference the timestamp of the current and previous events. The event-based \textit{Corner Detector} module finds relevant features from the events triggered by dynamic objects. The \textit{Optical Flow} module determines the direction of motion of the corners belonging to moving objects. Optical flow is computed only from corner events to reduce the computational cost. Finally, \textit{Clustering} gathers optical flow measurements to compute the average flow of the detected dynamic objects. Algorithm \ref{alg:obstacle_detection} describes the proposed dynamic obstacle motion estimation method. 
Previous event-based methods for dynamic obstacle detection rely either on optimization methods \cite{mitrokhin2018event} or  motion compensation techniques \cite{falanga2020dynamic} to distinguish events belonging to moving objects. 
Our approach focuses on detecting events from dynamic objects performing low computational processing and low latency response suitable for \textit{sense-and-avoid} onboard ornithopters. The \textit{Time Filter} module detects events generated from moving objects using as reference the timestamp difference of the events triggered at the same pixel location. The surface of active events (SAE) $S \in \mathbb{R}^2$ maps the event coordinates $\mathbf{x}$ with the timestamp $ts$ of the last occurring event at $S(\mathbf{x})$. Thus, $S$ describes a 2D representation of the timestamp evolution of triggered events. Under the assumption that objects move with a high relative velocity w.r.t. the robot, the events triggered by dynamic objects are detected using a time threshold $\tau$. If the time difference $\Delta ts$ between an incoming event $e_k$ and $S(\mathbf{x}_k)$ is lower than $\tau$ the event is considered to belong to a moving object. Next, the timestamp of $e_k$ updates $S$ by $S(\mathbf{x}_k) = ts_k$ for the future evaluations.
The value of $\tau$ depends on the velocity of the robot, defined by $\tau = \alpha \tau_{v_{z}} + (1 - \alpha) \tau_{\omega_{x}}$, where $\tau_{v_z}$ and $\tau_{\omega_x}$ are the contributions due to the current linear forward and pitch angular velocities of the robot ($v_z$ and $\omega_x$, respectively), and $\alpha \in [0, 1]$ sets the contribution of each velocity component. $\tau_{v_{z}}$ and $\tau_{\omega_{x}}$ are defined as follows:  
\begin{equation}
\scalemath{0.95}{
\begin{bmatrix}
\tau_{v_{z}} \\ \tau_{\omega_{x}}
\end{bmatrix}
=
\begin{bmatrix}
\tau_{H} \\ \tau_{H}
\end{bmatrix}
-
(\tau_{H}-\tau_{L})
\begin{bmatrix}
(v_z - v_{L})(v_{H} - v_{L})^{-1} \\ (\omega_x - \omega_{L})(\omega_{H} - \omega_{L})^{-1}
\end{bmatrix},
}
\label{eq:tau}
\end{equation}
\noindent
where $[v_{L}$, $v_{H}]$ and $[\omega_{L}, \omega_{H}]$ are the typical velocity ranges of the robot flight. The operation range $[\tau_L, \tau_H]$ is empirically selected, see Section \ref{sec:experimental}. 

\vspace{-0.3cm}
\begin{algorithm}[h!]
\SetAlgoLined
\KwIn{$\textbf{e}_k(\textbf{x}_k, ts_k, p)$}
\KwOut{$\boldsymbol{(\overline{v_x},\overline{v_y})}$}
  
  \If(\Comment{\small{\textit{Time Filter} evaluation}}){$ts_k - S(\mathbf{x}_k)$ $< \tau$}{
        updateCurrentSlice($\textbf{e}_k$) \\
        \If(\Comment{\small{\textit{Corner Detector}}}){isCorner($\textbf{e}_k$)}{
        $(v_x, v_y)$ $\xleftarrow{}$ computeOpticalFlow() \\
        $id \leftarrow{}$updateClusters($\textbf{e}_k, v_x, v_y$) \\
        $(\overline{v_x},\overline{v_y}) \leftarrow{}$ getClusterFlow($id$)
        }
        updateClusters($ts_k$)  \Comment{\small{Retrieve obstacle flow}}
    }
  $S(\mathbf{x}_k) = ts_k$  \Comment{\small{Update time reference at $\mathbf{x}_k$}}\\   
  $d,\tau_c \leftarrow{}$ sliceRotation($\textbf{e}_k$)\\
 \caption{Event-based moving obstacle detection}
 \label{alg:obstacle_detection}
\end{algorithm}
\vspace{-0.5cm}

Next, the events belonging to dynamic obstacles are processed to estimate their direction of motion, module \textit{Optical Flow} in Fig. \ref{fig:obstacle_detection_diagram}. That optical flow provides the relative motion estimation between the camera and the objects. Our approach uses the event-based optical flow method ABMOF \cite{liu2018adaptive}. It is  
based on block matching operations between event slices. Slices are 2D histograms of events collected during the accumulation time $d$. ABMOF includes two different control strategies to vary $d$ depending on the event generation through time. Despite using slices of accumulated events to compute optical flow, it performs \textit{event-by-event} processing by computing the flow of each incoming event. Our method integrates an adapted version of ABMOF to reduce the onboard processing.
First, event slices are fed only with events triggered from moving objects. This reduces the computational load by processing only $<$$32\%$ of the event stream in the experiments of Section \ref{sec:experimental}. Second, the optical flow is computed only from events considered as corners. The *eFast event \textit{Corner Detector} in \cite{scaramuzza2017fast} is selected for this task given its fast response and low False Positive Rate. Computing optical flow only from corners reduces the computational cost in $>$$25 \%$ while providing a stable optical flow estimation as described in Section \ref{sec:experimental}.

Finally, the resulting event optical flow estimations are clustered to obtain an approximation of the object's optical flow. For this task, we used an adapted version of the \textit{event-by-event} clustering algorithm described in \cite{rodriguez2020async}. This algorithm clusters events with spatio-temporal continuity within an adaptive time window. The algorithm was modified to cluster optical flow from events. It receives as input the tuple $f = (\mathbf{x},ts,\mathbf{v})$, where $\mathbf{v} = (v_x,v_y)$ represents the optical flow estimation of the event with pixel coordinates $\mathbf{x}$  and timestamp $ts$. Clustering is performed by evaluating the proximity of each new event to a randomly selected event from each of the previous clusters. Each cluster is defined by its centroid $\overline{\mathbf{x}}$, which represents the average location of cluster events, the average optical flow $\overline{\mathbf{v}} =  (\overline{v_x},\overline{v_y})$, and the list $\Phi$ of previous tuples $f$ assigned to the cluster. Each new valid optical flow estimation updates $\overline{\mathbf{v}}$ as in Eq. \ref{eq:centroid_add}.

\begin{minipage}{0.47\linewidth}
    \begin{equation}
    \scalemath{0.9}{
    \overline{\mathbf{v}} := \frac{\eta}{\eta+1}\overline{\mathbf{v}} + \frac{1}{\eta+1}\mathbf{v},
    \label{eq:centroid_add}
    }
    \end{equation}
\end{minipage}
\begin{minipage}{0.47\linewidth}
    \begin{equation}
    \scalemath{0.9}{
    \overline{\mathbf{v}} :=    \frac{\eta}{\eta-1}\overline{\mathbf{v}} - \frac{1}{\eta-1}\mathbf{v}_{\dagger},
    \label{eq:centroid_sub}
    }
  \end{equation}
\end{minipage}
\vspace{0.05 cm}

\noindent
where $\eta$ is the number of assigned tuples to the cluster (i.e., the length of $\Phi$). Similarly, the influence of old samples is removed from $\overline{\mathbf{v}}$, see Eq. \ref{eq:centroid_sub},
where $\mathbf{v}_{\dagger}$ corresponds to flow samples with timestamps lower than $\tau_c$. The parameter $\tau_c$ defines the maximum lifetime of flow samples in the cluster and it is dynamically adjusted using the reference feedback from \cite{liu2018adaptive}. Finally, 
ASAP is used to prevent processing overflows by dynamically adapting event packaging to keep the responsiveness of the method.

\subsection{Collision Risk Evaluation Strategy}
\label{subsec:evasive_policy}

\begin{figure}[b!]
     \vspace{-0.3cm}
	\centering
	\includegraphics[width=.9\linewidth]{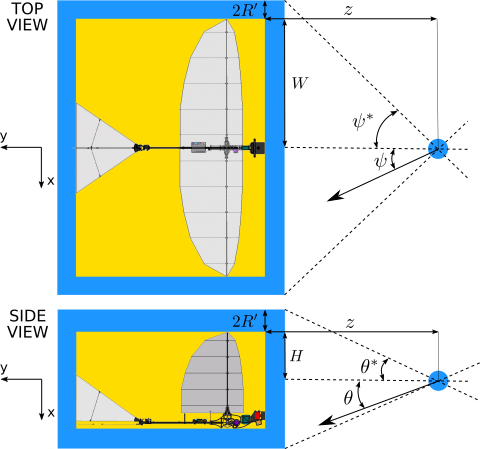}
	\caption{Collision risk evaluation based on geometry constraints. 
	$\psi^*$ and $\theta^*$ are used to determinate collision risk.}
	\label{fig:collision}
\end{figure}

A reactive evasive maneuver strategy is used to prevent collision risk situations with incoming obstacles. The possible collisions with detected obstacles are determined by considering the geometry of the robot. The ornithopter is approximated by a $2W \times 2H$ safety volume and the obstacle is approximated by a sphere of radius $R'$ (see Fig. \ref{fig:collision}). $R'$ encloses the obstacle volume and a \textit{Safety Distance} $b$ considering small sensing uncertainties, i.e.,  $R'$=$R$$+$$b$.
Thus, the minimum angles $\psi$ and $\theta$ to avoid a possible collision risk without deviating the flight trajectory are:
\begin{equation}
\scalemath{0.93}{
    \psi^{\ast}(t) = \arctan{\frac{W+2R'}{z(t)-2R'}},
    \quad
    \theta^{\ast}(t) = \arctan{\frac{H+2R'}{z(t)-2R'}},
}
\label{eq:angle_limits}
\end{equation}

\noindent
where $z(t)$ is the obstacle depth w.r.t. the camera. For any pair of angles ($\psi$, $\theta$) such that $|\psi(t_1)| \leq |\psi^{\ast}(t_1)|$ and $|\theta(t_1)| \leq |\theta^{\ast}(t_1)|$ a possible collision between the robot and the obstacle might occur at $t \geq t_1$ if the robot trajectory is not modified. In these cases, an evasive maneuver is activated by guiding the ornithopter away from the obstacle trajectory. The collision evaluation of Eq. \ref{eq:angle_limits} depends of the robot-obstacle depth $z(t)$ and its size. Three evaluation cases are considered based on the obstacle available information:

\begin{itemize}
    \item Assuming $z(t)$ is directly measurable (e.g., using a time-of-flight sensor) and $R$ is known, the collision risk evaluation is directly implemented using Eq. \ref{eq:angle_limits}.
    \item Assuming only $R$ is known, $z(t)$ can be estimated by $z(t) = \lambda R/L(t)$, where $\lambda$ is the camera focal length, and $L(t)$ is the largest side of a rectangle enclosing the clustered events. The value of $L(t)$ is computed by analysing the spatial distribution of events in the cluster. 
    \item If neither $z(t)$ nor $R$ are known, the collision risk cannot be predicted. Thus, any detected obstacle triggers an evasive maneuver.
\end{itemize}

The second case, having prior information of the object geometry to perform collision risk evaluation, is experimentally validated in Section \ref{sec:experimental}. 
We adopt a conservative strategy that detects collision risk situations using geometrical considerations and exerts reactive evasion maneuvers using the information of the obstacle motion. This reduces processing requirements, enabling low-latency execution.

\subsection{Tail Control}
Flapping-wing robots present more complex dynamics than multirotors and fixed-wing robots. Their non-holonomic underactuated nature requires considering the robot dynamic restrictions to evaluate the effect of the actuation commands in future spatial configurations. Conversely, reactive obstacle avoidance requires a fast and robust response to perform aggressive movements. Evasion maneuvers must be performed as fast as possible, thus the control method must be computationally simple. Additionally, since the robot is solely controlled using onboard perception --which implies high levels of uncertainty, obstacle avoidance control must be robust to minimize their effect.
One feasible control strategy is to perform an avoidance maneuver with an opposite direction velocity vector w.r.t. the incoming obstacle velocity. From the obstacle mean optical flow $\overline{\mathbf{v}}$ estimated in Section \ref{subsec:obstacle_detection}, the controller computes the longitudinal, $\delta_e$, and lateral, $\delta_r$, tail deflections to perform the evasive maneuver. To accomplish both robust and computationally efficient requirements, the following control law is implemented:

\begin{equation}
    \mathbf{u_{react}} = -(\boldsymbol{\kappa_0} + \boldsymbol{\kappa_1} \|\overline{\mathbf{v}}\|) \circ \overline{\mathbf{v}},
    \label{eq:control}
\end{equation}

\noindent
where $\circ$ denotes the Hadamard product, $\mathbf{u_{react}} = [\delta_e^{react}, \delta_r^{react}]^T$ is the control action, and $\boldsymbol{\kappa}_0$ and $\boldsymbol{\kappa}_1$ are controller gains. The values of $\boldsymbol{\kappa}_0$ and $\boldsymbol{\kappa}_1$ were experimentally tuned. The followed criterion was to achieve fast control response when an obstacle is detected.
The previous solution assumes that the best maneuver to avoid an obstacle is to fly in the opposite direction to its velocity. However, this strategy is agnostic to the robot dynamics. To cope with this limitation, our method also considers a flapping-wing linearized model adapted from \cite{armanini2016time} to compute the tail deflections. The model parameters were initially approximated empirically and then fitted by performing different tests inside a motion capture system. The ornithopter linear and angular velocities and its attitude are estimated using an onboard inertial navigation system.
From the current robot configuration, and given $\mathbf{u_{react}}$ from Eq. \ref{eq:control}, the adapted model is used to compute the robot acceleration for each tail angle in a discretized range between $\delta_e^{react} \pm 10^\circ$ for longitudinal deflection and between $\delta_r^{react} \pm 10^\circ$ for lateral deflection. The selected tail angle increments, $\mathbf{\Delta u_{model}} = [\Delta \delta_e^{model}, \Delta \delta_r^{model}]^T$, are those that give the robot the greatest acceleration --i.e., those that increase the speed of the evasion maneuver the most in the shortest time. Hence, the tail deflections to command are composed as $\mathbf{u} = \mathbf{u_{react}} + \mathbf{\Delta u_{model}}$. Additionally, our method constrains the deflection values considering the restrictions presented in \cite{guzman2021design} to avoid stall.

\section{Experiments}
\label{sec:experimental}
The experimental platform is the \textit{E-Flap} robot, a customized ornithopter developed at the GRVC Robotics laboratory. The robot has a total length of 95 cm, a wingspan of 1.5 m, a weight of 510 g, and a maximum payload of 520 g at the expense of limited maneuverability and reduced flight time. The sensors and hardware are placed along the robot body to improve the platform maneuverability. A low-cost Khadas VIM3 handles onboard online perception and control. It equips a VectorNav VN-200 inertial navigation system that provides measurements of the ornithopter's body frame velocity. A DAVIS346 event camera provides low latency polarized events from its integrated dynamic vision sensor DVS.  The camera weight was reduced to a third of its original value to satisfy the robot weight restrictions. The camera mounts a lens with a total weight of $5$ g, a Field of View of 68$^{\circ}$ horizontal and  53.5$^{\circ}$ vertical, and an IR-Cut filter to cope with the IR emissions from the motion capture system. The event-based obstacle detector method, the evasive maneuver strategy and the flapping-wing controller avoidance were implemented in C++ using \textit{ROS}. In all the experiments, ASAP was configured to provide event packages at $250$ Hz to cope with the low computational restrictions. Thus, the mean optical flow was updated each $4$ ms. It is worth mentioning that the \textit{event-by-event} property of our algorithm allows configuring the optical flow update by changing the rate of the packages provided by ASAP. 

\begin{figure}[t]
    \begin{center}
        \includegraphics[width=0.85\linewidth]{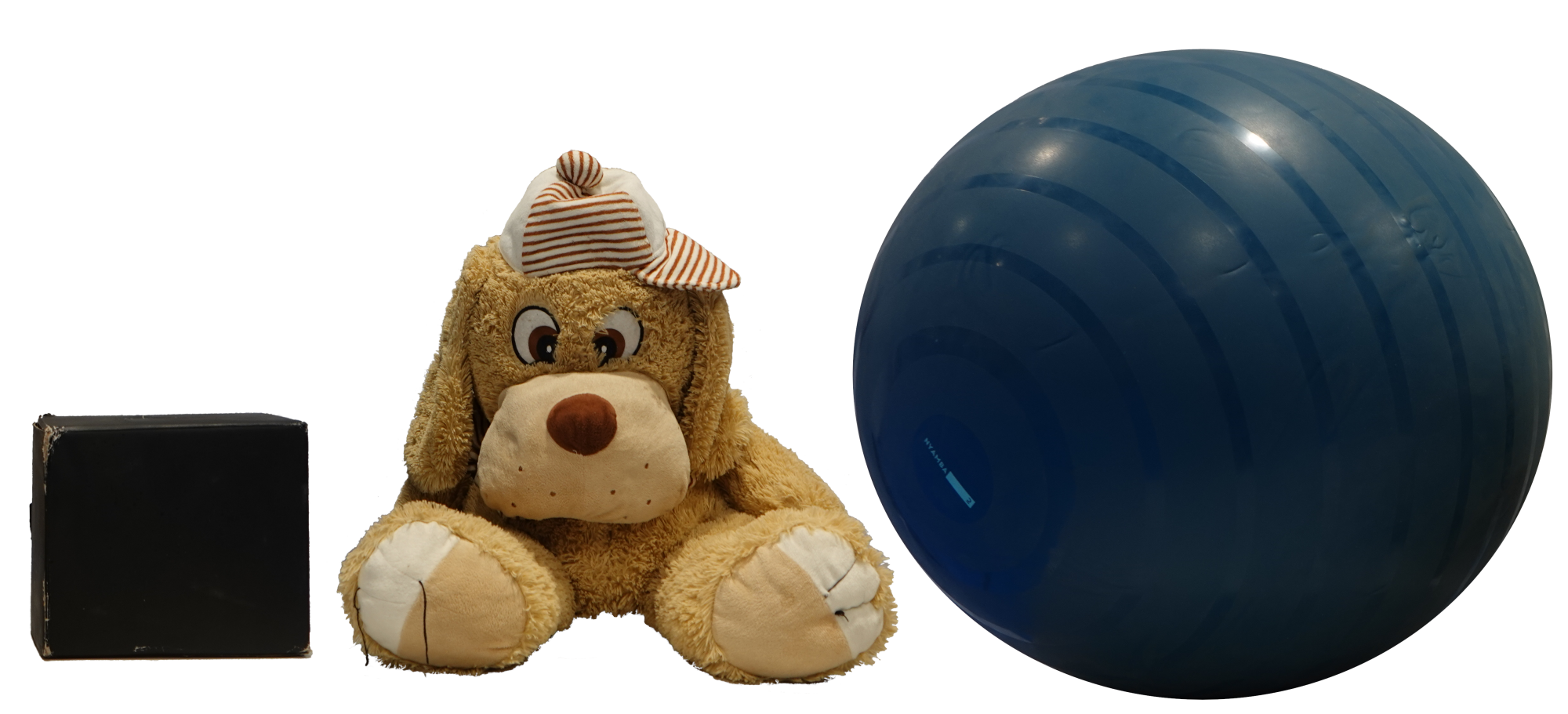}
        \caption{Different size objects considered for the experiments: left) Small Box; center) Stuffed Toy; and right) FitBall.}
        \label{fig:obstacles}
    \end{center}
    \vspace{-0.7cm}
\end{figure}

The proposed dynamic obstacle \textit{sense-and-avoid} system was experimentally validated in both indoor and outdoor scenarios. The GRVC Testbed is a closed area of $15 \times 21 \times 8$ m with $24$ \textit{OptiTrack Prime$^x$ $13$} cameras providing millimeter accuracy pose estimations. 
The outdoor scenario is a Soccer Field of $48 \times 54$ m with surrounding obstacles suitable to perform short flight experiments with the ornithopter.
For all the experiments, the parameters in Eq.\ref{eq:tau} were set to $v_L$=$3$ m/s, $v_H$=$6$ m/s, $\omega_L$=$1.3$ rad/s, and $\omega_H$=$3$ rad/s, given by the typical flight kinematics conditions of the robot. Further, $\alpha$ was set to $0.8$, as the robot mainly performs forward motions. Parameter $\tau$ sets the threshold to distinguish the events triggered by dynamic objects. A large value entails detections at longer distances while permitting events triggered by static objects. Besides, a small value of $\tau$ performs a quite selective filtering at the expense of reducing the distance at which the obstacles are detected. Hence, $\tau$ defines a trade-off between filtering events triggered by static objects and the maximum distance to detect an obstacle. From our experiments, $\tau_L$=$15$ ms and $\tau_H$=$25$ ms were empirically selected to provide a trade-off between having detection distances of $6$ m while filtering $82\%$ of events triggered by the scene background. 

The experimental validation was divided into two parts. First, the dynamic obstacle detection and motion estimation were evaluated. Second, the evaluation of the full obstacle avoidance system was performed in experiments in indoor and outdoor scenarios with different illumination conditions.

\subsection{Obstacle detection and motion estimation evaluation}

\begin{figure*} [h!]
\begin{center}
\begin{minipage}{0.3\linewidth}
\includegraphics[width=1\linewidth]{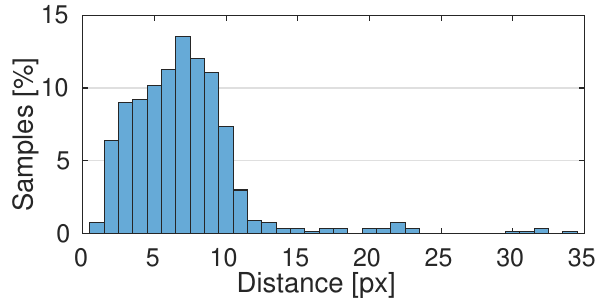}
\vspace{-3.5ex}
\captionof{figure}{Histogram of the distance between the detected object and the Ground Truth in the overall tests. }
\label{fig:histogram}
\end{minipage}
\hspace{1pt}
\begin{minipage}{0.68\linewidth}
\renewcommand{\tabcolsep}{2.6pt}
\renewcommand{\arraystretch}{1.2}
\begin{center}
\scalebox{0.86}{
\begin{tabular}{c|c|c|c|c|c|c|c|c|c|c|c|c|}
\cline{2-13}
   & \multicolumn{4}{c|}{$\mathbf{1.5 - 2.0}$ \textbf{m}} & \multicolumn{4}{c|}{$\mathbf{2.0 - 4.0}$ \textbf{m}} & \multicolumn{4}{c|}{$\mathbf{4.0 - 6.0}$ \textbf{m}} \\ \cline{2-13} 
   & \textit{Acc} &  \textit{Pre} & \textit{TPR}  & \textit{FPR}  &  \textit{Acc} &  \textit{Pre} & \textit{TPR}  & \textit{FPR}  & \textit{Acc} & \textit{Pre} & \textit{TPR}  & \textit{FPR}  \\ \hline
\multicolumn{1}{|c|}{\textbf{Small Box}} & 0.91 & 0.95 & 0.86 & 0.04 & 0.97 & 0.98 & 0.97 & 0.03 & 0.80 & 0.86 & 0.75 & 0.09 \\ \hline
\multicolumn{1}{|c|}{\textbf{Stuffed Toy}} & 0.97 & 0.95 & 0.97 & 0.04 & 0.93 & 0.98 & 0.86 & 0.02 & 0.92 & 0.97 & 0.85 & 0.02 \\ \hline
\multicolumn{1}{|c|}{\textbf{Fitball}} & 0.91 & 0.98 & 0.82 & 0.01 & 0.94 & 0.94 & 0.93 & 0.05 & 0.92 & 0.94 & 0.91 & 0.08 \\ \hline
\end{tabular}}
\captionof{table}{Accuracy, Precision, True Positive Rate (TPR), and False Positive Rate (FPR) results of dynamic obstacle detection using the different obstacles while varying the launching distance.}
\label{tab:obstacle_detection}
\end{center}
\end{minipage}
\end{center}
\vspace{-2.5em}
\end{figure*}

\begin{figure*}[b]
    \vspace{-0.6cm}
    \begin{center}
        \includegraphics[width=0.45\linewidth]{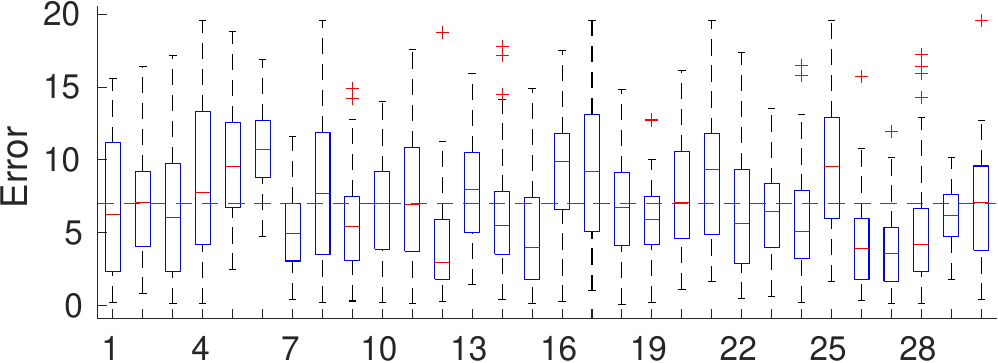}
        \includegraphics[width=0.52\linewidth]{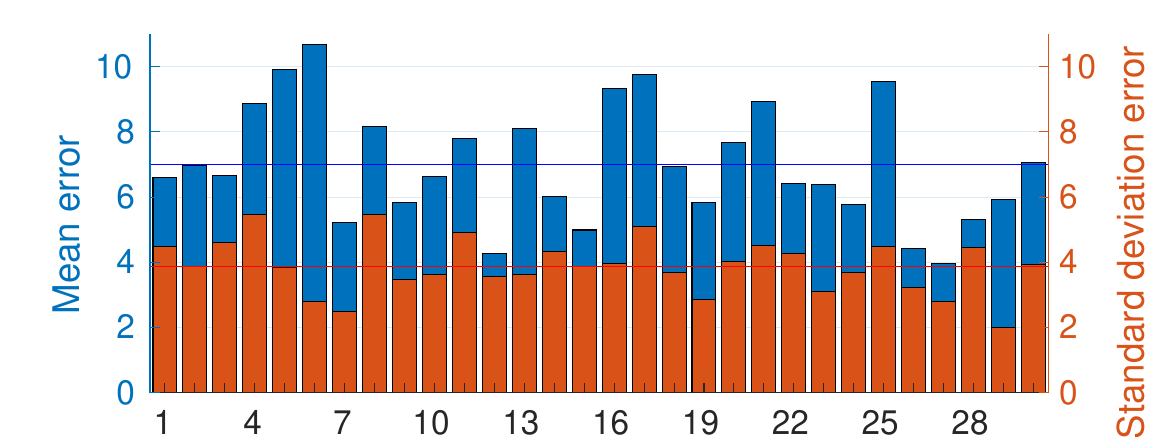}
        \caption{
        Box plot (left) and mean and standard deviation (right) of the motion direction error along each of 30 experiments. The direction error is defined as the instantaneous absolute difference between the direction of motion estimated by our approach and the obstacle direction from the ground truth. The mean error considering all the experiments is $6.97^\circ$. The mean standard deviation considering all the experiments is $3.89^\circ$, and maximum instantaneous direction error is $19.54^\circ$.
        }
        \label{fig:motion_estimation}
    \end{center}
\end{figure*}

In these experiments, obstacles were thrown into the Field of View (FoV) of the camera while the ornithopter performed forward flight. The experiments were performed in the Testbed scenario to retrieve the pose information from the obstacle and the robot. The goal was to evaluate the detection and motion estimations performance of the method described in Section \ref{subsec:obstacle_detection}. Three objects with different sizes were considered (Fig. \ref{fig:obstacles}): a Small Box of size 220 $\times$ 200 $\times$ 150 mm, a Stuffed Toy of size 400 $\times$ 450 $\times$ 400 mm, and a Fitball with a diameter of 750 mm. Obstacle detection was evaluated in distances ranging from $0.5$ m to $6$ m. At distances closer than 0.5 m the obstacle body filled a large zone of the image leading to invalid detections. At large distances obstacles were represented by few pixel events due to the camera resolution, which hindered their 2D representation for obstacle detection. A total of 45 experiments were performed with each obstacle. 

The detection performance was evaluated by comparing the outcome of our algorithm with the frames provided by the APS sensor. The centroid of the detected moving object was rendered into \textit{event images} obtained each $25$ ms. The distance between the object's centroid in both frames was used as evaluation criteria. Each comparison led to a possible result:  False Positive (FP), False Negative (FN), True Positive (TP), and True Negative (TN). A True Positive occurred when the Manhattan distance between the centroid of both objects was $\leq$$\eta$ pixels. Fig. \ref{fig:histogram} shows a histogram of the distance between the object and the ground of truth in the performed experiments. The majority of samples at distances $>$$15$ px correspond to False Positives. Choosing $\eta$ to $10$ was a suitable trade-off between validating the majority valid samples as True Positives while rejecting False Positive samples.
Next, the overall Accuracy, Precision, True Positive Rate (TPR), and False Positive Rate (FPR) were computed. Table \ref{tab:obstacle_detection} summarizes the detection results of each obstacle at different distances. The method reported an overall accuracy of $91.7 \%$. The distance directly affected the detection performance specially with small objects at distances $>$$4$ m. At such distances, small objects triggered few events hampering object detection, and the lack of triggered events affected the obstacle detection as many events in $S(\mathbf{x})$ did not satisfy the $\tau$ condition. In general, at distances between $2$ to $4$ m the method reported an accuracy above $94.7 \%$. In this range, the size of the different objects in the image was enough to perform successful dynamic object detection in the majority of the experiments. Additionally, the method provided a low number of False Positives as reported by the average FPR of $4.2 \%$, and an average Precision of $95.1 \%$. Finally, a few detections were missed during the experiments as evidenced by the average TPR of $88.5 \%$. 

The obstacle detection method was evaluated in multi-obstacle experiments in which three objects were thrown from different directions in the camera FoV at similar times. It reported an accuracy of $74.2\%$ among all the experiments. The performance reduction was caused by the generation of False Positive samples when the obstacles overlapped in the image plane, which tended to merge clusters hampering the individual detection of objects. Despite this degradation, the method results were satisfactory taking into account that the detection was performed with a single event camera.

The motion estimation evaluation consisted of comparing the mean optical flow direction of the dynamic obstacle on the image plane with the ground truth direction. The ground truth was obtained from the motion capture system by projecting the position of the obstacle in the image and estimating its motion direction from previous and future samples. For better validation, the obstacles were launched from a distance of $8$ m to describe larger trajectories. A total of $30$ experiments were performed. The error was defined as the instantaneous angle difference between the estimated direction of the obstacle movement and the ground truth direction. Fig. \ref{fig:motion_estimation} shows the quartiles, mean, and standard deviation of the error along each experiment.
The resulting Root Mean Square Error was $11.2^{\circ}$, which is a reasonable error to guide the robot in a collision-free direction. \textit{Safety Distance} $b$ described in Section \ref{subsec:evasive_policy} was used to consider this error by enlarging the obstacle geometry to add extra safety to the evasion maneuver.

\subsection{\textit{Sense-and-avoid} evaluation}
\label{subsec:obstacle_avoidance}

Next, the full proposed \textit{sense-and-avoid} system was evaluated. Two cases can be distinguished in case the obstacle detection method fails. A False Positive obstacle detection triggers an unnecessary evasion maneuver. A False Negative obstacle detection neglects the evaluation of a collision risk situation, potentially causing an impact between the robot and the obstacle. Different sets of experiments were analysed to evaluate its performance. The ornithopter was launched towards a goal zone by an operator and performed forward flight using the controller described in \cite{maldonado2020adaptive}. One of the obstacles in Fig. \ref{fig:obstacles} was launched to intercept the ornithopter. The obstacles were launched in various directions to evaluate different evasive maneuvers. To enhance repeatability the lightweight obstacles were launched using a motorized launcher which set their initial speed to a specific value. Obstacles were thrown from a distance of $10$ m and moved with an average speed in the range of $[5, 8]$ m/s. Our system checked for collision risk situations and activated an avoidance maneuver if a collision risk situation was detected. Three types of analyses were performed.

First, we performed experiments to analyse the performance of the system in case an \textit{Intersection of the Safety Volumes (ISV)} occurred. An \textit{ISV} exists when the robot safety volume and the sphere of radius $R'$ enclosing the object intersect along the robot trajectory in case it performs no evasion maneuver. These experiments were performed indoors: the ground truth robot and obstacle positions were recorded using a motion caption system to check if an \textit{ISV} occurred. A total of $25$ experiments were performed with each object using regular ($760$ lx) and dark ($<$$15$ lx) illumination conditions. The average avoidance success rate with the three objects, see Table \ref{tab:oa_evaluation}, was $90.7\%$. The best result was obtained using the FitBall as obstacle, which larger size allowed an earlier detection of the collision risk situation. The experiments with dark illumination conditions also reported a remarkable success rate: the slight  performance degradation was caused by the additional noisy events produced by the poor lighting.

Second, we performed experiments to analyze the system performance in case \textit{ISV} situations did not occur. 20 experiments were performed using the Small Box obstacle. In $85\%$ of the experiments, the system did not detect collision risk situations. Only in $15\%$ of the cases, it detected collision risk situations --activating an unnecessary evasive maneuver-- in flights with no \textit{ISV} situations. This result was mainly caused by the conservative selection of the radius $R'$ which enlarged the obstacle size to reduce impacts with the robot body. False Positive detections increased in the dark lighting experiments due to the higher level of noisy events.

Finally, the system performance in outdoor experiments was analyzed to evaluate its robustness to different scenarios. In these experiments, the collisions were evaluated visually due to the lack of motion capture information. A total of $25$ experiments were performed with each object under light and dark lighting conditions. The results are shown in Table \ref{tab:oa_evaluation}. The proposed system had a success rate of $92.0\%$ with regular illumination conditions, and reported acceptable results with dark lighting conditions $85.3\%$.

\begin{table}[h]
\begin{tabular}{p{1.0cm}|p{1.25cm}|p{1.25cm}|p{1.4cm}|p{0.75cm}|}
\cline{2-5}
 & \textbf{Validation}  & \textbf{Small Box} & \textbf{Stuffed Toy} & \textbf{FitBall} \\ \hline
\multicolumn{1}{|c|}{\textbf{Indoors}}       & \multirow{2}{*}{\makecell{\hspace{-0.2cm} Motion \\ \hspace{-0.2cm} Capture Sys.} }& \hfil 92\%   & \hfil 92\%   & \hfil 96\%  \\ \cline{1-1} \cline{3-5}  
\multicolumn{1}{|c|}{\textbf{Indoors Dark}}  &   & \hfil 84\%  & \hfil 88\%   & \hfil 92\% \\ \hline
\multicolumn{1}{|c|}{\textbf{Outdoors}}      & \multirow{2}{*}{ \hspace{0.1cm} Visual }    & \hfil 92\% & \hfil  92\%  & \hfil 92\% \\ \cline{1-1} \cline{3-5} 
\multicolumn{1}{|c|}{\textbf{Outdoors Dark}} &   & \hfil 84\%  & \hfil 84\%  & \hfil 88\%  \\ \hline
\end{tabular}
\caption{Success rate of the proposed dynamic obstacle avoidance method in different scenarios.}
\label{tab:oa_evaluation}
\end{table}

\section{Conclusions and Future Work}
\label{sec:conclusion}
This paper presents the first event-based obstacle avoidance system for large-scale flapping-wing robots. The proposed approach exploits the advantages of event-based vision to detect dynamic obstacles and perform evasion maneuvers. Our scheme has been validated in several indoor and outdoor scenarios with different illumination conditions. It reports an average avoidance success rate of $89.7\%$ evading dynamic obstacles of different sizes and shapes. Further, its \textit{event-by-event} processing nature and efficient implementation allow fast onboard computation even in low processing capacity hardware, providing high rate estimations ($250$ Hz in our experiments). Future work includes the validation of our method in other agile robots. Further, the development of a map-based method for the avoidance of static obstacles and its integration in a complete obstacle avoidance system for static and dynamic objects are object of future research.

\bibliographystyle{IEEEtran}
\bibliography{references}

\end{document}